\documentclass[a4]{article}
\usepackage{hyperref}
\usepackage{url}
\usepackage[T1]{fontenc}
\usepackage[utf8x]{inputenc}
\usepackage{amsmath}
\usepackage{empheq}
\usepackage{amsfonts}
\usepackage{amssymb}
\usepackage{amsthm}
\usepackage{graphicx, textcomp}
\usepackage{pst-math}
\usepackage{pst-node}
\usepackage{subfig}
\usepackage{float}
\usepackage[affil-it]{authblk}

\title{Predictable Feature Analysis}

\author{Stefan~Richthofer\footnote{Electronic address: \texttt{stefan.richthofer@ini.rub.de}; Corresponding author}, Laurenz~Wiskott\footnote{Electronic address: \texttt{laurenz.wiskott@ini.rub.de}}}
\affil{Institute For Neural Computation,\\ Ruhr-Universit\"at Bochum, Germany}

\providecommand{\abs}[1]{\lvert#1\rvert}

\providecommand{\norm}[1]{\lVert#1\rVert}

\providecommand{\av}[1]{\left\langle#1\right\rangle}

\providecommand{\bigav}[1]{\big\langle#1\big\rangle}

\providecommand{\coloneq}[0]{\mathrel{\mathop:}=}
\providecommand{\id}[0]{\mathbf{I}}
\providecommand{\trph}[0]{\Omega_t}

\DeclareMathOperator{\err}{err}

\DeclareMathOperator*{\eq}{=}
\DeclareMathOperator*{\appr}{\approx}

\DeclareMathOperator{\mvec}{vec}
\DeclareMathOperator{\hist}{hist}
\DeclareMathOperator{\diag}{diag}
\DeclareMathOperator{\tr}{Tr}

\DeclareMathOperator{\orth}{O}

\DeclareMathOperator*{\opmin}{minimize}

\providecommand{\optmin}[1]{\displaystyle\opmin_{#1} \qquad}
\DeclareMathOperator*{\subjectto}{subject \; to \qquad}
\providecommand{\pfaerr}[2]{\err(#1, #2)}

\providecommand{\pfaerrob}[2]{\bigav{\; \norm{#2-#1}^2 \; }}
\providecommand{\pfaerrr}[3]{\av{\; \norm{#2-#1(#3)}^2 \; }}

\providecommand{\pfaerrrb}[3]{\bigav{\; \norm{#2-#1(#3)}^2 \; }}

\newtheoremstyle{defi}
  {}
  {}
  {\slshape}
  {}
  {\bfseries}
  {}
  {\newline}
  {}

\theoremstyle{defi}

\newtheorem{definition}{Definition}
\newtheorem{lem}{Lemma}
\newtheorem{thm}{Theorem}

\begin{document}

\maketitle

\begin{abstract}
Every organism in an environment, whether biological, robotic or virtual,
must be able to predict certain aspects of its environment in order to survive
or perform whatever task is intended.
It needs a model that is capable of estimating the consequences of possible actions, so that planning, control, and decision-making become feasible.
For scientific purposes, such models are usually created in a problem specific manner using differential equations and other techniques from control- and system-theory.
In contrast to that, we aim for an unsupervised approach that builds up the desired model in a self-organized fashion. Inspired by Slow Feature Analysis (SFA), our approach is to extract sub-signals from the input, that behave as predictable as possible. These “predictable features” are highly relevant for modeling, because predictability is a desired property of the needed consequence-estimating model by definition. In our approach, we measure predictability with respect to a certain prediction model.
We focus here on the solution of the arising optimization problem and present a tractable algorithm based on algebraic methods which we call Predictable Feature Analysis (PFA).
We prove that the algorithm finds the globally optimal signal, if this signal can be predicted with low error.
To deal with cases where the optimal signal has a significant prediction error, we provide a robust, heuristically motivated variant of the algorithm and verify it empirically.
Additionally, we give formal criteria a prediction-model must meet to be suitable for measuring predictability in the PFA setting and also provide a suitable default-model along with a formal proof that it meets these criteria.
\end{abstract}

\pagebreak

\section{Introduction}
The motivation for Predictable Feature Analysis (PFA) comes from
typical reinforcement-learning settings, where an autonomous agent is placed in
an environment and aims to achieve some goal. While many common scenarios
are discrete (board-game-like) with rather few states, we consider more natural
scenarios, where the input is a continuous signal over
time and of high dimension like vision or some other sensory input.\footnote[1]{
Since it comes from a technical setup, the signal would still be discrete.
However, we would not regard its discreteness as states but would
conceptually treat it as a continuous signal.}

PFA is intended as a tool to help the agent make sense of this vast amount of
incoming data. Our approach is to look for information in the signal that
helps to understand and manipulate the environment in the desired way. To achieve
its goal, the agent must be able to plan its actions and thus needs to
understand, how the environment behaves -- it needs a model that is capable
of predicting the outcomes of possible actions. It has been frequently proposed that predictable
features are crucial to obtain such a model, see \cite{BialekNemenmanEtAl-2001} for a review.
In contrast to most common approaches
from control theory, we attempt to perform the modeling without putting previously known,
problem-specific information (usually a representation of the environment in
form of differential equations and system theoretic setups) into the model,
but look for a truly unsupervised, self-organized approach.

Slow Feature Analysis (SFA) is an algorithm that has most characteristics we are
looking for (and as such also served as the name-giving pattern for PFA). It is an
algorithm that has proven valuable in several fields and problems concerning
signal- and data analysis. The idea is that a drastic, yet reasonable dimensionality
reduction can be obtained by focusing on slowly varying sub-signals, the so-called
“slow features”. These are considered most relevant, because slowness usually
indicates invariance and invariant problem representations are crucial for typical
data-analysis and recognition tasks, such as regression and classification. Many of these
tasks have proven to become much more feasible on the reduced signal after SFA has been
applied. For instance, tasks like the self-organization of complex-cell receptive fields, the recognition of whole objects invariant to spatial transformations, the self-organization of
place-cells, extraction of driving forces, or nonlinear blind source separation were successfully
performed on the basis of SFA (see \cite{DahneWilbertEtAl-2010a, FranziusWilbertEtAl-2011, FranziusSprekelerEtAl-2007d, Wiskott-2003b, BlaschkeZitoEtAl-2007}).
%

PFA extracts sub-signals from the input using the same
methods like SFA does, but instead of the slowest features, it selects those that are best
predictable by a certain prediction model. In section \ref{sec:criteria} we give the criteria a model
must meet to be suitable for this purpose. While there are also model-independent notions of predictability like in the information bottleneck approach \cite{CreutzigGlobersonEtAl-2009, CreutzigSprekeler-2008}, focusing on concrete models has the advantage that if PFA finds predictable sub-signals, one is directly able to actually perform the prediction, since the appropriate model is given.
The arising optimization problem turned out to
be significantly harder than that of SFA, because it is a nested problem: The features extracted must be optimized for predictability, but judging their predictability is an optimization problem by itself.
Optimizing these problems in turns
usually converges to sub-optimal solutions that depend on the starting point. In this work
we discuss the details of the PFA-problem and present a tractable algorithm, thus setting up the
basis for future PFA-related work.

There have been other approaches to use notions of predictability.
For instance \cite{gepperth:simultaneous} considers scenarios involving embodied agents. 
They let two sensors predict each other in order to retrieve representation-invariant information.
\cite{CreutzigSprekeler-2008} combines notions of predictability with SFA to better understand principles of sensory coding strategies.
There also exists an ICA-based approach to predictability-driven dimensionality reduction, see \cite{Hyvarinen01}.
ForeCA (Forecastable Component Analysis), an independently developed method, is based on the same paradigm as PFA, but proposes a model-independent approach \cite{goerg13}. A further difference is that PFA (optionally) searches for well predictable Systems, while ForeCA selects best predictable single components. In future work, we are going to compare PFA- and ForeCA-results to get a better understanding for strengths and weaknesses of the two approaches. Finally, there has also been a previous version of our PFA approach \cite{RichthoferWeghenkelWiskott-2012c}.

\section{Extracting predictable features} \label{sec:extraction}
Given an input-signal $\mathbf{x}(t)$ with $n$ components, our goal is to extract a certain number ($r$) of well predictable output-components, referred to as “predictable features”. Since our approach is inspired by SFA, we start with a summary of that algorithm.

\subsection{Recall SFA} \label{sec:sfa}
In the SFA-setting, the optimized property is slow variation. Extraction is in principle performed by linear transformation and projection.\footnote[2]{In a strict sense, the transformation is affine because it clears the signal's mean. Additionally one can count the non-linear expansion as part of the extraction.} The parameters of these mappings are optimized over a finite training-phase $\trph$ consisting of equidistant time points. To make the method more powerful, a non-linear expansion $\mathbf{h}$ can be applied to the signal -- usually using monomials of low degree.

In order to avoid the trivial constant solution, the output is constrained to have unit variance and zero mean. Additionally, the output-components must be pairwise uncorrelated. This way the repeated occurrence of the same component is avoided.
Mean is defined using $\av{s(t)}_t~\coloneq~\frac{1}{\abs{\trph}} \sum_{t\in \trph} s(t)$ (average of a signal over the training phase). To fulfill the constraints, the expanded signal is sphered over the training-phase, i.e. its mean is shifted to zero and the covariance-matrix is normalized to the identity matrix:
\begin{align} \label{sphering}
\tilde{\mathbf{z}}(t) \quad &\coloneq \quad \mathbf{h}(\mathbf{x}(t)) - \av{\mathbf{h}(\mathbf{x}(t))}_t &&\text{(make mean-free)}\\
\mathbf{z}(t) \quad &\coloneq \quad \mathbf{S} \tilde{\mathbf{z}}(t) \qquad \text{with} \quad \mathbf{S} \coloneq \av{\tilde{\mathbf{z}} \tilde{\mathbf{z}}^T}^{-\frac{1}{2}} &&\text{(normalize covariance)}
\end{align}
Summing up all SFA-constraints, the following optimization problem is derived:
\begin{align} \label{sfa}
\text{For} \; i \in \{1, \ldots, r\} \notag \\
	\begin{split} 
		\optmin{\mathbf{a}_i \in \mathbb{R}^{n}} & \mathbf{a}_i^T \av{\dot{\mathbf{z}}\dot{\mathbf{z}}^T} \mathbf{a}_i\\
		\subjectto	& \mathbf{a}_i^T \av{\mathbf{z}} \hphantom{\mathbf{a}_i \mathbf{a}_j \mathbf{z}^T} \, = \quad 0 \quad \hphantom{\forall \; j < i} \quad \text{(zero mean)}\\
				& \mathbf{a}_i^T \av{\mathbf{z} \mathbf{z}^T} \mathbf{a}_i \hphantom{\mathbf{a}_j} = \quad 1 \quad \hphantom{\forall \; j < i} \quad \text{(unit variance)}\\
				& \mathbf{a}_i^T \av{\mathbf{z} \mathbf{z}^T} \mathbf{a}_j \hphantom{\mathbf{a}_i} = \quad  0 \quad \forall \; j < i \quad \text{(pairwise uncorrelated)}
	\end{split}
\end{align}
To describe the algorithm, we first define the extraction matrix $\mathbf{A}_r \coloneq \left(\mathbf{a}_1, \ldots, \mathbf{a}_r \right) \in \mathbb{R}^{n \times r}$ and the reduced identity $\mathbf{I}_r \in \mathbb{R}^{n \times r}$ consisting of the first $r$ euclidean unit vectors as columns.
Now please note that because of the sphering, it holds that $\av{\mathbf{z}} = 0$ and $\av{\mathbf{z} \mathbf{z}^T} = \id$, thus having the constraints equal to
\begin{equation}
\exists \mathbf{A} \in \orth(n) \colon \qquad \mathbf{A}_r \quad = \quad \mathbf{A}\mathbf{I}_r
\end{equation}
where $\orth(n) \subset \mathbb{R}^{n \times n}$ denotes the space of orthogonal transformations, i.e. $\mathbf{A}\mathbf{A}^T = \mathbf{I}$.
Choosing $\mathbf{a}_i$ as the eigenvectors of $\av{\dot{\mathbf{z}}\dot{\mathbf{z}}^T}$, corresponding to the eigenvalues in
descending order, yields an $\mathbf{A}_r$ that solves \eqref{sfa} globally. We denote the extracted signal with $\mathbf{m} \coloneq \mathbf{A}_r^T \mathbf{z}$. \cite{WiskottBerkesEtAl-2011} describes this procedure in detail.

\subsection{Modeling the PFA-problem} \label{sec:pfa}
In order to measure predictability, we focus on a certain prediction-model.
Because it is simple and very popular, we use \textbf{linear, auto-regressive prediction} as our default model -- it is successfully used in many fields for modeling time-related problems.
A signal is regarded well predictable, if each value can be approximated by a linear combination of some ($p$) recent values.
Expressing this formally, we face the problem of finding vectors $\mathbf{a}$ and $\mathbf{b}$ such that
\begin{align}
 \mathbf{a}^T \mathbf{z}(t) \quad \appr^! \;& \quad b_{1} \mathbf{a}^T \mathbf{z}(t-1) + \ldots + b_{p} \mathbf{a}^T \mathbf{z}(t-p) \label{pfa-criterionARScalar}\\
= \;& \quad \mathbf{a}^T \hist_{\mathbf{z}, p}(t) \; \mathbf{b}
\end{align}
with $\hist$ defined as the signal's history over the recent $p$ time-steps:
\begin{equation}
\label{history}
	\hist_{\mathbf{z}, p, \Delta}(t) \quad \coloneq \quad \sum_{i=1}^p \quad \mathbf{z}(t-i\Delta) \mathbf{e}_i^T \quad \text{with} \quad \mathbf{e}_i \in \mathbb{R}^p, \quad \left(\mathbf{e}_1, \ldots, \mathbf{e}_p \right) = \mathbf{I}_{p, p}.
\end{equation}
Here $\mathbf{I}_{p, p}$ denotes the $p$-dimensional identity, thus $\mathbf{e}_i$ denotes the $i$-th $p$-dimensional Euclidean unit vector. 
$\Delta$ defaults to $1$: $\hist_{\mathbf{z}, p} \coloneq \hist_{\mathbf{z}, p, 1}$.

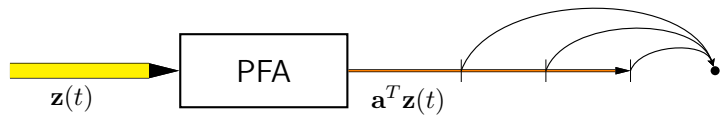
\begin{figure}[ht]
   \centering
   \subfloat{

\psset{xunit=1.5,yunit=1.0}
\begin{pspicture} (0.25, 1)(7, 3)
\sffamily

\psline[doubleline=true, doublesep=0.19, doublecolor=yellow, linewidth=0.01, arrowsize=0.2, arrowlength=2, arrowinset=0]{->}(0.5, 2)(2, 2)
\psframe(2, 1.5)(3.5, 2.5)
\rput[c](2.75, 2){\large{PFA}}
\rput[t](1.05, 1.8){$\mathbf{z}(t)$}
\psline[doubleline=true, doublesep=0.02, doublecolor=orange, linewidth=0.01, arrowsize=0.1, arrowlength=2, arrowinset=0]{->}(3.5, 2)(6, 2)
\rput[tl](3.7, 1.8){$\mathbf{a}^T \mathbf{z}(t)$}

\pnode(4.5, 2){A}
\pnode(5.25, 2){B}
\pnode(6, 2){C}
\cnode*(6.75, 2){0.06}{D}

\psline[linewidth=0.01, linestyle=dashed](4.5, 1.85)(4.5, 2.15)
\psline[linewidth=0.01, linestyle=dashed](5.25, 1.85)(5.25, 2.15)
\psline[linewidth=0.01, linestyle=dashed](6, 1.85)(6, 2.15)

\ncarc[linewidth=0.015, arcangle=70, arrowsize=0.1, arrowlength=1, arrowinset=0.2]{->}{C}{D}
\ncarc[linewidth=0.015, arcangle=70, arrowsize=0.1, arrowlength=1, arrowinset=0.2]{->}{B}{D}
\ncarc[linewidth=0.015, arcangle=70, arrowsize=0.1, arrowlength=1, arrowinset=0.2]{->}{A}{D}
\end{pspicture}

%
%
%
%

   }
   \caption{Illustration of PFA}
   \label{fig:PFAIllustration}
 \end{figure}

Like in SFA, we optimize the parameters over the training phase $\trph$ and also adopt the SFA constraints to avoid trivial or repeated solutions. The first steps of PFA are indeed equal to those in SFA, i.e. we also allow for a non-linear expansion and also start with a sphering-step. As far as possible, we use the notation that was introduced in \ref{sec:sfa}.
The common way to extend \eqref{pfa-criterionARScalar} to multiple dimensions can be written as
\begin{equation} \label{pfa-criterionARDiagMatrix}
 \mathbf{m}(t) \quad \appr^! \quad \mathbf{B}_{1} \mathbf{m}(t-1) + \ldots + \mathbf{B}_{p} \mathbf{m}(t-p) \quad \text{with}\quad \mathbf{B}_i \in \mathbb{R}^{n \times n} \text{, diagonal}
\end{equation}
In this form, it does not fulfill all criteria from section \ref{sec:criteria} (it is not orthogonal agnostic for $n > 1$). Nevertheless, we mention strategies to solve \eqref{pfa-criterionARDiagMatrix} in the appendix, section \ref{sec:extractIsolated}. Here we proceed by refining it to be suitable for PFA:
\begin{equation} \label{pfa-criterionARGeneralMatrix}
 \mathbf{m}(t) \quad \appr^! \quad \mathbf{B}_{1} \mathbf{m}(t-1) + \ldots + \mathbf{B}_{p} \mathbf{m}(t-p) \quad \text{with}\quad \mathbf{B}_i \in \mathbb{R}^{n \times n} \hphantom{\text{, diagonal}}.
\end{equation}
The difference to the first formulation is that each extracted component's prediction may depend on all other extracted components. Note that \eqref{pfa-criterionARDiagMatrix} and \eqref{pfa-criterionARGeneralMatrix} are equal for $n = 1$. 
A massive advantage of model \eqref{pfa-criterionARGeneralMatrix} is that we can initially fit it to our data in full dimension and search for the best-fitted components afterwards. For \eqref{pfa-criterionARDiagMatrix}, this would not be possible, because the fitting-quality of each component is not invariant under the transformation used for extraction. We formalize the need for such a model-property in section \ref{sec:criteria}.

To formalize fitting, we briefly introduce the notion of a general prediction model and of the fitting-error. We speak of a general prediction-model 
$\mathbf{g}$ as

\begin{equation} \label{generalPredictionModel}
	\mathbf{g} \in \mathcal{G}: \qquad \mathbf{z}(t) \quad \appr^! \quad \mathbf{g}(\hist_{\mathbf{z}, p, \Delta}(t))
\end{equation}
where $\mathcal{G}$ is the model-class, i.e. the set of possible realizations of $\mathbf{g}$.
We measure the prediction error in an average least squares sense by
\begin{align}
	&\err(\mathbf{g}, \mathbf{z}) 
	\quad \coloneq \quad \pfaerrr{\mathbf{g}}{ \mathbf{z}}{ \hist_{\mathbf{z}, p, \Delta}} \label{totalPredictionError}
\end{align}

Now fitting a prediction-model on a given sample in a least squares sense can be expressed as
\begin{equation} \label{fitting_g0}
	\optmin{\mathbf{g} \in \mathcal{G}} \pfaerr{\mathbf{g}}{\mathbf{z}} 
\end{equation}
By $\mathbf{g}_{\mathbf{z}}^*$, we denote the global solution of \eqref{fitting_g0} and define the following shortcut-notation:
\begin{align}
	&\err(\mathbf{z}) \quad \coloneq \quad \pfaerr{\mathbf{g}^*_\mathbf{z}}{ \mathbf{z}} \label{shortcutError}
\end{align}
To formalize \eqref{pfa-criterionARGeneralMatrix} as a prediction model in this notation, we combine the coefficient-matrices $\mathbf{B}_{i}$ to a single broad matrix and define $\mathbf{g}_{\mathbf{B}}$ and $\mathcal{G}_{\eqref{pfa-criterionARGeneralMatrix}}$:
\begin{align}
 \mathbf{B}
 \hphantom{\mathbf{g}_{\mathbf{B}}(\hist_{\mathbf{z}, p}(t))\mathcal{G}_{\eqref{pfa-criterionARGeneralMatrix}}}
 &\coloneq \qquad \left( \mathbf{B}_{1}, \ldots, \mathbf{B}_{p} \right) \quad \in \; \mathbb{R}^{n \times np} \label{B} \\
 \mathbf{g}_{\mathbf{B}}(\hist_{\mathbf{z}, p}(t))
 \hphantom{\mathbf{B} \mathcal{G}_{\eqref{pfa-criterionARGeneralMatrix}}}
 &\coloneq \qquad \mathbf{B} \mvec(\hist_{\mathbf{z}, p}(t)) \label{gB} \\
 \mathcal{G}_{\eqref{pfa-criterionARGeneralMatrix}}
 \hphantom{\mathbf{g}_{\mathbf{B}}(\hist_{\mathbf{z}, p}(t))\mathbf{B}} 
 &\coloneq \qquad \lbrace \; \mathbf{g}_{\mathbf{B}} \colon \quad \mathbf{B} \in \; \mathbb{R}^{n \times np} \; \rbrace 
\end{align}
By analytic optimization, we obtain the following regression formula to fit $\mathbf{g}_{\mathbf{B}} \in \mathcal{G}_{\eqref{pfa-criterionARGeneralMatrix}}$ to $\mathbf{m}~=~\mathbf{A}_r^T \mathbf{z}$:
\begin{equation} \label{B-ReducedDimRankTransformSolution}
 \mathbf{B}_{\mathbf{z}}(\mathbf{A}_r) \quad = \quad \mathbf{A}^T_r \av{\mathbf{z}\mathbf{\zeta}^T}\underline{\mathbf{A}_r} \left(\underline{\mathbf{A}_r}^T\av{\mathbf{\zeta}\mathbf{\zeta}^T} \underline{\mathbf{A}_r} \right)^{-1}
\end{equation}
Here we used $\zeta(t) \coloneq \mvec(\hist_{\mathbf{z}, p}(t))$ and the following shortcut notation defined for any matrix $\mathbf{M}$:
\begin{equation} \label{multiA}
\underline{\mathbf{M}} \quad \coloneq \quad \mathbf{I}_{p, p} \otimes \mathbf{M} \quad = \qquad \underbrace{\!\!\!\!\!\!\left( \begin{smallmatrix} \mathbf{M}&  & \mathbf{0} \\  & \ddots &  \\ \mathbf{0} &  & \mathbf{M} \end{smallmatrix} \right)\!\!\!\!\!\!}_{\text{$p$ times $\mathbf{M}$}} 
\end{equation}
If $r=n$, $\mathbf{A}=\mathbf{I}$ and thus $\mathbf{A}_r=\mathbf{I}$, we write
$\mathbf{W} \coloneq \mathbf{B}_{\mathbf{z}}(\mathbf{I}) = \av{\mathbf{z}\mathbf{\zeta}^T} \av{\mathbf{\zeta}\mathbf{\zeta}^T}^{-1}$.
(See \ref{sec:notation} for an overview of all notation in this document.)
It sometimes happens that $\av{\mathbf{\zeta}\mathbf{\zeta}^T}$ is not (cleanly) invertible due to some very small or even zero eigenvalues. We regard it best practice to project away the eigenspaces corresponding to eigenvalues below a critical threshold. The intuition behind this is that spaces corresponding to (almost-)zero eigenvalues indicate redundancies in the signal and should not be used for prediction anyway. To perform this, first compute an eigenvalue decomposition on $\av{\mathbf{\zeta}\mathbf{\zeta}^T}$. Replace eigenvalues below the threshold by $0$ and replace the other ones by their multiplicative inverse. After that undo the decomposition and use the resulting matrix as a proxy for $\av{\mathbf{\zeta}\mathbf{\zeta}^T}^{-1}$.

For $r=n$ and $\mathbf{A} \in \orth(n)$, we have $\mathbf{B}_{\mathbf{z}}(\mathbf{A}) = \mathbf{A}^T\mathbf{W}\underline{\mathbf{A}}$. Since $\mathbf{z}$ is sphered, we can state the following compact notation of the PFA-problem:
\begin{equation} \label{pfaAutoRegressiveWhitened}
\optmin{\mathbf{A} \in \orth(n)} \;
\err(\mathbf{A}_r^T \mathbf{z}) 
\end{equation}
Inserting our default model, we have $\err(\mathbf{A}_r^T \mathbf{z}) = \pfaerrob{\mathbf{B}_{\mathbf{z}}(\mathbf{A}_r) \underline{\mathbf{A}_r}^T \zeta}{ \mathbf{A}_r^T \mathbf{z}}$.
However, because \eqref{B-ReducedDimRankTransformSolution} is an involved term, mainly due to the projection under an inversion symbol, \eqref{pfaAutoRegressiveWhitened} appears to be intractable by every method known to us\footnote[3]{Not counting evolutionary and other inherent local optimization approaches, since we aim for the global solution. Experiments showed us that locally optimal solutions are usually still of high error and of low relevance for the model.}. Instead of solving it directly, we propose the following tractable relaxation:
\begin{equation} \label{pfaNonAutoRegressiveWhitened}
\optmin{\mathbf{A} \in \orth(n)} \; \pfaerrob{\mathbf{I}_r^T\mathbf{B}_{\mathbf{z}}(\mathbf{A}) \underline{\mathbf{A}}^T \zeta}{ \mathbf{A}_r^T \mathbf{z}} \quad = \quad \bigav{\; \norm{\mathbf{A}_r^T(\mathbf{z}-\mathbf{W}\mathbf{\zeta})}^2 \; }
\end{equation}

Informally speaking, problem \eqref{pfaNonAutoRegressiveWhitened} asks for components that are optimally predictable, if the prediction
may be based on the entire input signal, rather than just on the extracted components themselves.
From now on we denote a global optimum of \eqref{pfaAutoRegressiveWhitened} with $\mathbf{A}_r^*$ and of \eqref{pfaNonAutoRegressiveWhitened} with $\mathbf{A}_r^{(0)}$.

\begin{figure}[ht]
   \centering
   \subfloat{

\psset{xunit=1.5,yunit=1.0}
\begin{pspicture}  (0.25, 1)(7, 3)
\sffamily

\psline[doubleline=true, doublesep=0.19, doublecolor=yellow, linewidth=0.01](0.5, 1.75)(2, 1.75)
\psline[doubleline=true, doublesep=0.19, doublecolor=yellow, linewidth=0.01, linestyle=dashed](2, 1.75)(3.5, 1.75)
\psline[doubleline=true, doublesep=0.19, doublecolor=yellow, linewidth=0.01, arrowsize=0.2, arrowlength=2, arrowinset=0]{->}(3.5, 1.75)(6, 1.75)
\psframe(2, 1.25)(3.5, 3)
\rput[c](2.75, 2.5){\large{PFA}}
\rput[b](1.25, 2){$\mathbf{z}(t)$}
\psline[doubleline=true, doublesep=0.02, doublecolor=orange, linewidth=0.01, arrowsize=0.1, arrowlength=2, arrowinset=0]{->}(3.5, 2.5)(6, 2.5)
\rput[bl](3.7, 2.6){$\mathbf{m}(t) = \mathbf{A}_r^T \mathbf{z}(t)$}

\pnode(4.5, 1.75){A}
\pnode(5.25, 1.75){B}
\pnode(6, 1.75){C}
\cnode*(6.75, 2.5){0.06}{D}

\psline[linewidth=0.01, linestyle=dashed](4.5, 1.5)(4.5, 2.6)
\psline[linewidth=0.01, linestyle=dashed](5.25, 1.5)(5.25, 2.6)
\psline[linewidth=0.01, linestyle=dashed](6, 1.5)(6, 2.6)

\ncarc[linewidth=0.015, arcangle=10, arrowsize=0.1, arrowlength=1, arrowinset=0.2]{->}{C}{D}
\ncarc[linewidth=0.015, arcangle=-7, arrowsize=0.1, arrowlength=1, arrowinset=0.2]{->}{B}{D}
\ncarc[linewidth=0.015, arcangle=-10, arrowsize=0.1, arrowlength=1, arrowinset=0.2]{->}{A}{D}
\end{pspicture}
}
\caption{Illustration of relaxation \eqref{pfaNonAutoRegressiveWhitened}}
\end{figure}
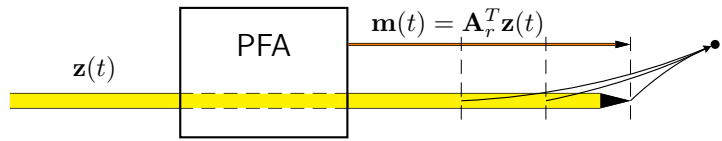
To solve \eqref{pfaNonAutoRegressiveWhitened} globally, we write it as
\begin{equation} \label{pfaNonAutoRegressiveWhitenedRewrite}
 \optmin{\mathbf{A} \in \orth(n)} \; \tr \left( \mathbf{A}_r^T \bigav{\left(\mathbf{z} - \mathbf{W}\mathbf{\zeta}\right) \left(\mathbf{z} - \mathbf{W}\mathbf{\zeta}\right)^T} \mathbf{A}_r \right)
\end{equation}
and choose $\mathbf{A}$ such that it diagonalizes
$\bigav{\left(\mathbf{z} - \mathbf{W}\mathbf{\zeta}\right) \left(\mathbf{z} - \mathbf{W}\mathbf{\zeta}\right)^T}$
and sorts the $r$ smallest eigenvalues to the upper left. This method can be described as performing PCA on the residuals of the least squares fit. By some calculus, one can show formal equivalence of this approach to the method proposed in \cite{boxTiao1977}. In section \ref{sec:proofOfRelaxationGap} we prove that if $\err((\mathbf{A}_r^*)^T \mathbf{z}) = 0$, then $\mathbf{A}_r^{(0)}$ is also a global solution of
\eqref{pfaAutoRegressiveWhitened}.\footnote[4]{Note that if $\mathbf{A}_r^{(0)}$ is used as solution for \eqref{pfaAutoRegressiveWhitened}, the prediction model must be refitted to the reduced signal to get optimal prediction. For this, calculate $\mathbf{B}_{\mathbf{z}}(\mathbf{A}_r^{(0)})$ as defined in \eqref{B-ReducedDimRankTransformSolution}.} 
More precisely speaking, the relaxation gap of \eqref{pfaNonAutoRegressiveWhitened} depends on $\err((\mathbf{A}_r^*)^T \mathbf{z})$ in a continuous manner and is zero, if that error is zero. If the optimal sub-signal has a significant prediction error, the solution obtained as $\mathbf{A}_r^{(0)}$ usually suffers from overfitting and is sub-optimal for \eqref{pfaAutoRegressiveWhitened}. In the following, we offer a heuristic method to overcome this overfitting.

\subsection{Avoiding overfitting} \label{sec:overfitting}
To reduce overfitting, we propose the heuristics that signals well predictable in terms of \eqref{pfaAutoRegressiveWhitened} yield a lower error-propagation to subsequent predictions than signals that are well predictable in terms of \eqref{pfaNonAutoRegressiveWhitened} but not in terms of \eqref{pfaAutoRegressiveWhitened}. We ground this on the intuition that the prediction of the latter ones is partly based on noisy data -- thus subsequent predictions inherit a higher error.
To formalize this idea we define
\begin{equation} \label{V}
\mathbf{V} \quad \coloneq \quad \av{\mathbf{\zeta}(t+1)\mathbf{\zeta}^T(t)}_t \av{\mathbf{\zeta}\mathbf{\zeta}^T}^{-1}
\end{equation}
and can thus perform iterated prediction as follows:
\begin{equation} \label{zVi}
\mathbf{z}(t) \quad \approx \quad \mathbf{W}\mathbf{V}^i \mathbf{\zeta}(t-i)
\end{equation}

\begin{figure}[ht]
   \centering
   \captionsetup[subfloat]{labelformat=empty}
   \subfloat{

\psset{xunit=1.5, yunit=1.0}
\begin{pspicture}  (0.25, 1)(7.1, 3)
\sffamily

\psline[doubleline=true, doublesep=0.19, doublecolor=yellow, linewidth=0.01](0.5, 1.75)(2, 1.75)
\psline[doubleline=true, doublesep=0.19, doublecolor=yellow, linewidth=0.01, linestyle=dashed](2, 1.75)(3.5, 1.75)
\psline[doubleline=true, doublesep=0.19, doublecolor=yellow, linewidth=0.01, arrowsize=0.2, arrowlength=2, arrowinset=0]{->}(3.5, 1.75)(6, 1.75)
\psframe(2, 1.25)(3.5, 3)
\rput[c](2.75, 2.5){\large{PFA}}
\rput[b](1.25, 2){$\mathbf{z}(t)$}
\psline[doubleline=true, doublesep=0.02, doublecolor=orange, linewidth=0.01, arrowsize=0.1, arrowlength=2, arrowinset=0]{->}(3.5, 2.5)(6, 2.5)
\rput[bl](3.7, 2.6){$\mathbf{m}(t) = \mathbf{A}^T \mathbf{z}(t)$}

\pnode(4.5, 1.75){A}
\pnode(5.25, 1.75){B}
\pnode(6, 1.75){C}
\cnode*(6.75, 1.75){0.06}{D}
\cnode*(7.5, 1.75){0.06}{E}
\cnode*(8.25, 1.75){0.06}{F}
\cnode*(8.25, 2.5){0.06}{G}

\psline[linewidth=0.01, linestyle=dashed](4.5, 1.5)(4.5, 2.6)
\psline[linewidth=0.01, linestyle=dashed](5.25, 1.5)(5.25, 2.6)
\psline[linewidth=0.01, linestyle=dashed](6, 1.5)(6, 2.6)

\ncarc[linewidth=0.015, arcangle=40, arrowsize=0.1, arrowlength=1, arrowinset=0.2]{->}{C}{D}
\ncarc[linewidth=0.015, arcangle=40, arrowsize=0.1, arrowlength=1, arrowinset=0.2]{->}{B}{D}
\ncarc[linewidth=0.015, arcangle=40, arrowsize=0.1, arrowlength=1, arrowinset=0.2]{->}{A}{D}

\ncarc[linewidth=0.015, arcangle=40, arrowsize=0.1, arrowlength=1, arrowinset=0.2]{->}{D}{E}
\ncarc[linewidth=0.015, arcangle=40, arrowsize=0.1, arrowlength=1, arrowinset=0.2]{->}{C}{E}
\ncarc[linewidth=0.015, arcangle=40, arrowsize=0.1, arrowlength=1, arrowinset=0.2]{->}{B}{E}

\ncarc[linewidth=0.015, arcangle=40, arrowsize=0.1, arrowlength=1, arrowinset=0.2]{->}{E}{F}
\ncarc[linewidth=0.015, arcangle=40, arrowsize=0.1, arrowlength=1, arrowinset=0.2]{->}{D}{F}
\ncarc[linewidth=0.015, arcangle=40, arrowsize=0.1, arrowlength=1, arrowinset=0.2]{->}{C}{F}

\ncdiag[linewidth=0.015, angleA=90, angleB=-90, arrowsize=0.1, arrowlength=1, arrowinset=0.2]{->}{F}{G}
\end{pspicture}}
\caption{Illustration of iterated prediction}
\end{figure}
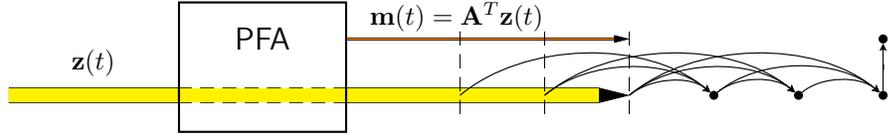

Based on this, we propose the following optimization problem:
\begin{equation} \label{pfaNonAutoRegressiveIterated}
\optmin{\mathbf{A} \in \orth(n)} \; \sum_{i=0}^k \quad \bigav{\; \norm{\mathbf{A}_r^T(\mathbf{z}-\mathbf{W}\mathbf{V}^i\mathbf{\zeta}(t-i)}^2 \; }_t
\end{equation}
We can solve it globally in a rather similar way like \eqref{pfaNonAutoRegressiveWhitened}. To do so, we write it as
\begin{equation} \label{pfaNonAutoRegressiveIteratedRewrite}
 \optmin{\mathbf{A} \in \orth(n)} \; \tr \Big( \mathbf{A}_r^T \sum_{i=0}^k \bigav{\left(\mathbf{z} - \mathbf{W}\mathbf{V}^i\mathbf{\zeta}(t-i)\right) \left(\mathbf{z} - \mathbf{W}\mathbf{V}^i\mathbf{\zeta}(t-i)\right)^T}_t \mathbf{A}_r \Big)
\end{equation}
and then solve it by diagonalizing $\sum_{i=0}^k \bigav{\left(\mathbf{z} - \mathbf{W}\mathbf{V}^i\mathbf{\zeta}(t-i)\right) \left(\mathbf{z} - \mathbf{W}\mathbf{V}^i\mathbf{\zeta}(t-i)\right)^T}_t$ and sorting the lowest $r$ eigenvalues to the upper left. We denote the global solution of \eqref{pfaNonAutoRegressiveIterated} by $\mathbf{A}^{(k)}_r$.
How to optimally choose $k$ for a certain problem is currently an open question, but we know from experiments that up to some value, increasing $k$ improves $\err((\mathbf{A}^{(k)}_r)^T \mathbf{z})$.
Beyond that value, increasing $k$ lowers the quality again. Our intuition is that the critical value is related to the maximal time distance, over which the signal holds any auto-correlation -- investigating this formally will be subject of future work. To give some impression of the technique and as a proof of concept, we demonstrate it on a synthetic example. We know that basic trigonometric functions are losslessly predictable with our default model for $p=2$. Because of the theorem in section \ref{sec:proofOfRelaxationGap}, it would make no sense to work with a losslessly predictable signal. So we add some white noise $\eta(t)$ to it and define an example signal $\mathbf{x}(t) \coloneq (\sin(0.1 t)+0.7 \eta(t), \sin(0.2 t)+\eta(t), \sin(0.4 t)+5.3 \eta(t))^T$. We train the algorithm with $\Delta = 1$ once on $1000$ samples and once on $2000$ samples, always extracting two components, i.e. $r=2, p=2$. As a lower bound for any error not involving overfitting, we evaluate \eqref{pfaNonAutoRegressiveWhitened}. For our example we get a lower bound of $\approx 1.206$ for $1000$ samples and $\approx 1.22$ for $2000$ samples. These are plotted as red horizontal lines in the following. To measure the amount of overfitting, we add many dimensions of white random data to our signal and mix everything up by a random, orthogonal transformation. While $\mathbf{x}$ has always the same noise-seed, the added data is generated with different noise in every run. The following results are averaged over about 150 runs, and plot the prediction error against the noise-dimension:\footnote[5]{Note that the vertical axis ranges from $1.15$ onward to provide a better focus.}
\begin{figure}[H]
\centering
\parbox{6cm}{
\includegraphics[width=7cm]{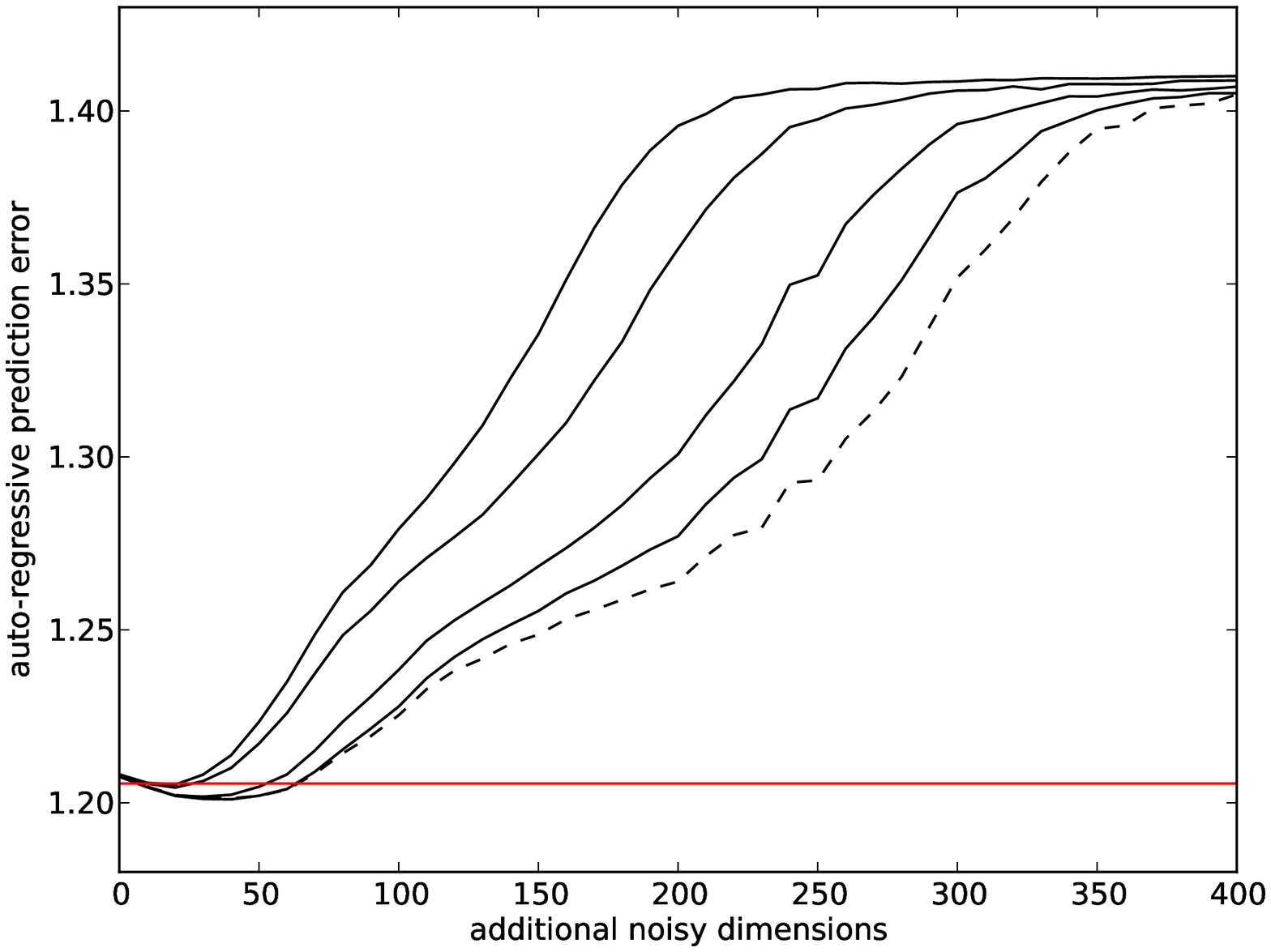}
\caption{$k=0,\ldots, 4$ from left to right$;$ $k=4$ dashed$;$ 1000 samples}
\label{fig:k0to4at1000}}
\quad
\begin{minipage}{6cm}
\includegraphics[width=7cm]{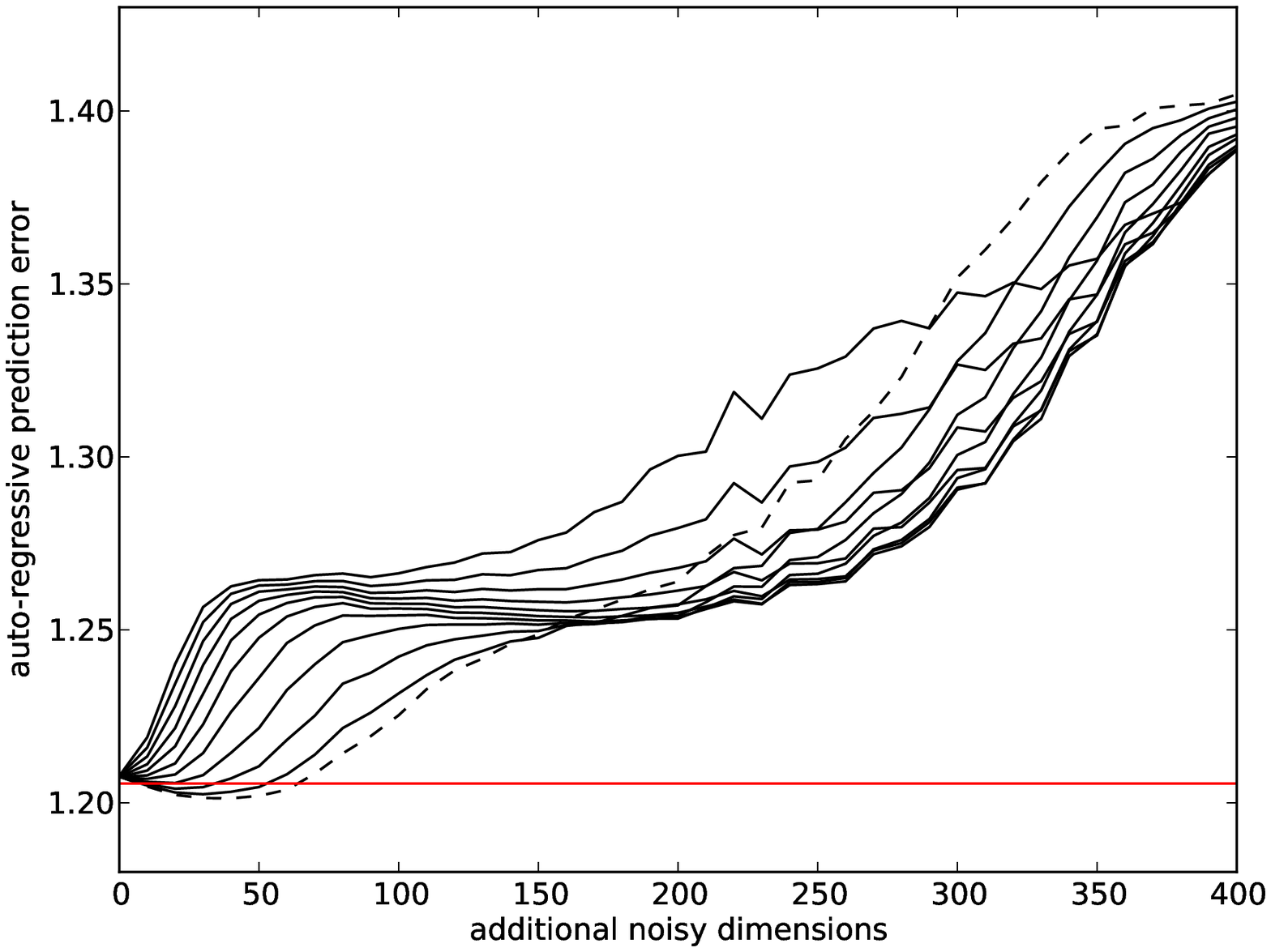}
\caption{$k=4,\ldots, 14$ from right to left$;$ $k=4$ dashed$;$ 1000 samples}
\label{fig:k4to14at1000}
\end{minipage}

\begin{minipage}{6cm}
\includegraphics[width=7cm]{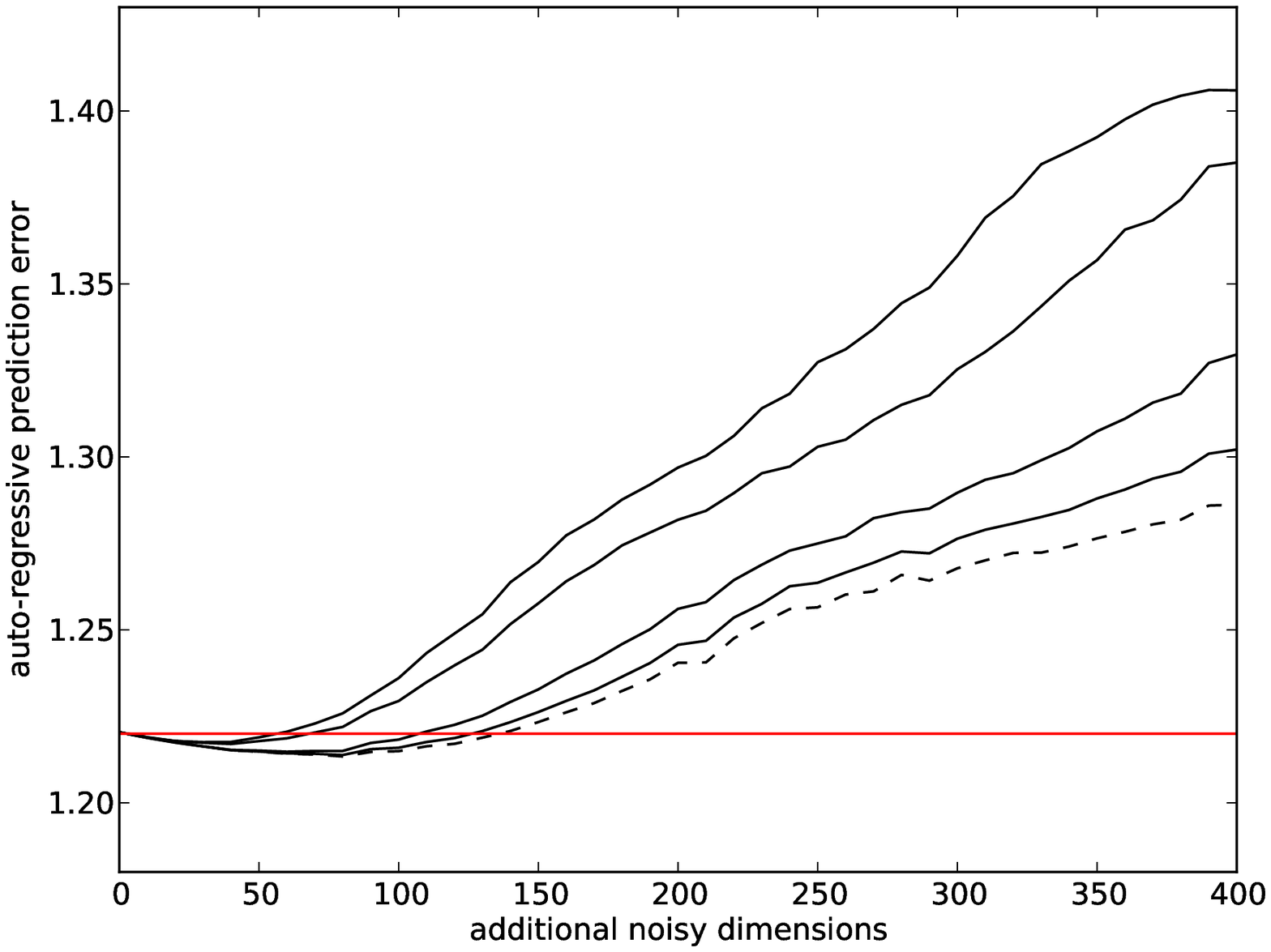}
\caption{$k=0,\ldots, 4$ from left to right$;$ $k=4$ dashed$;$ 2000 samples}
\label{fig:k0to4at2000}
\end{minipage}
\quad
\begin{minipage}{6cm}
\includegraphics[width=7cm]{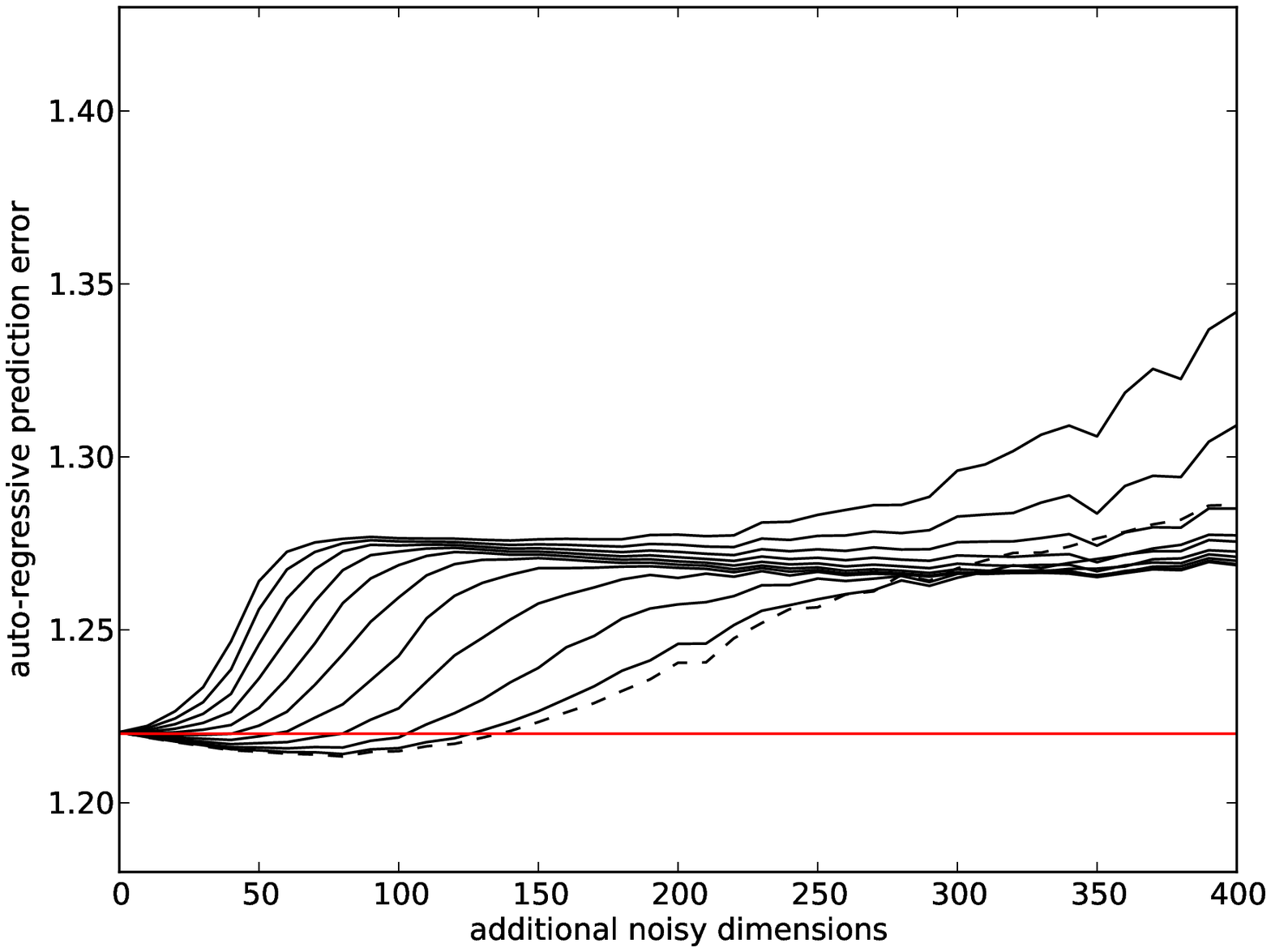}
\caption{$k=4,\ldots, 14$ from right to left$;$ $k=4$ dashed$;$ 2000 samples}
\label{fig:k4to14at2000}
\end{minipage}
\end{figure}
Obviously $k=4$ yields the best results in this example. The result for $k=0$ with no additional noise can be considered to be the optimal solution. Solutions below the red line indicate overfitting conform with \eqref{pfaAutoRegressiveWhitened}. This kind of overfitting can only be reduced by using larger samples. That it decreases for higher noise-dimension indicates that the best solution is not found any more. So we conclude that in our example, the algorithm is robust for about $50$ dimensional noise if $1000$ samples are used and for about $100$-dimensional noise, if $2000$ samples are used. Future work will include more detailed research about these relationships. 

\pagebreak 
\section{Criteria for suitable prediction models} \label{sec:criteria}
In this section we discuss what properties of a prediction model are crucial to make the procedure described in \ref{sec:pfa} feasible.
\begin{definition}[Orthogonal agnosticity criterion]
We say that a prediction-model $\mathcal{G}$ is \textbf{orthogonal-agnostic} on $\Omega_t$, if
for every $\mathbf{A} \in \orth(n), \mathbf{g} \in \mathcal{G}$:
\begin{equation} \label{orthogonalAgnosticity}
	\err(\mathbf{z}) \quad = \quad \err(\mathbf{A}^T \mathbf{z})
\end{equation}
\end{definition}
\eqref{orthogonalAgnosticity} means that the model fits equally well to any orthogonal transformation of the data.
In section \ref{sec:proofOfRelaxationGap} we will need a more restrictive variant of this criterion that additionally considers projections of the data to subspaces:
\begin{definition}[Projective orthogonal agnosticity criterion]\label{sec:projOrthAgn}
We say that a prediction-model $\mathcal{G}$ is \textbf{projective orthogonal-agnostic} on $\Omega_t$, if for every $\mathbf{A} \in \orth(n)$, $r \leq n$ the following holds:
\begin{equation} \label{projectiveOrthogonalAgnosticity}
	\pfaerrr{\mathbf{I}_r^T \mathbf{g}_{\mathbf{A}^T \mathbf{z}}^*}{\mathbf{A}_r^T \mathbf{z}}{ \mathbf{A}^T \hist_{\mathbf{z}, p}} \quad = \quad \pfaerrr{\mathbf{A}_r^T \mathbf{g}_{\mathbf{z}}^*}{\mathbf{A}_r^T \mathbf{z}}{\hist_{\mathbf{z}, p}}.
\end{equation}
\end{definition}
Note that for $r = n$, \eqref{projectiveOrthogonalAgnosticity} simplifies to \eqref{orthogonalAgnosticity} and projective orthogonal agnosticity becomes equivalent to ordinary orthogonal agnosticity (since the Frobenius-norm is invariant under orthogonal transformations).
An even stronger and very intuitive criterion is the following:
\begin{definition}[Commuting with orthogonal transformations]
We say that a prediction-model $\mathcal{G}$ \textbf{commutes with orthogonal transformations}, if for every $\mathbf{A} \in \orth(n)$ the following holds:
\begin{equation} \label{commutingWithOrthogonal}
	\mathbf{g}_{\mathbf{A}^T \mathbf{z}}^* \quad = \quad \mathbf{A}^T \mathbf{g}_{\mathbf{z}}^*.
\end{equation}
\end{definition}
It is rather obvious that this criterion implies projective and ordinary orthogonal agnosticity. To assure projective orthogonal agnosticity,
it is a straightforward procedure to construct models such that they commute
with orthogonal transformations.
\begin{definition}[Information consistency criterion]
We say that a prediction-model $\mathcal{G}$ is \textbf{information-consistent} on $\Omega_t$, if for every $\mathbf{A} \in \orth(n)$, $r \leq n$ the following holds:
\begin{equation} \label{informationConsistency}
	\pfaerrrb{\mathbf{g}_{\mathbf{A}_r^T \mathbf{z}}^*}{\mathbf{A}_r^T \mathbf{z}}{\mathbf{A}_r^T \hist_{\mathbf{z}, p}} \quad \geq \quad \pfaerrrb{\mathbf{A}_r^T \mathbf{g}_{\mathbf{z}}^*}{\mathbf{A}_r^T \mathbf{z}}{\hist_{\mathbf{z}, p}}
\end{equation}
\end{definition}
An information-consistent model always benefits from more data rather than getting confused by it.
Note that for $r = n$, \eqref{informationConsistency} follows from orthogonal agnosticity.
\begin{thm}
Model $\mathcal{G}_{\eqref{pfa-criterionARGeneralMatrix}}$ is projective orthogonal agnostic and information consistent.
\end{thm}
\begin{proof}
Projective orthogonal agnosticity follows, because the model commutes with orthogonal transformations, as $\mathbf{B}_{\mathbf{z}}(\mathbf{A}) = \mathbf{A}^T\mathbf{B}_{\mathbf{z}}(\mathbf{I})\underline{\mathbf{A}}$.
To show that $\mathcal{G}_{\eqref{pfa-criterionARGeneralMatrix}}$ is information consistent, we need the solution of the following optimization problem:
\begin{equation} \label{proj_fitting_g}
	\optmin{\mathbf{B} \in \mathbb{R}^{n \times np}} \pfaerrr{\mathbf{I}_r^T\mathbf{g}_{\mathbf{B}}}{\mathbf{A}_r^T\mathbf{z}}{\mathbf{A}^T \hist_{\mathbf{z}, p}}
\end{equation}
Analytically we find a (not unique) solution to be $\mathbf{B}_{\mathbf{z}}(\mathbf{A})$. 
Note that $\mathbf{B}_{\mathbf{z}}(\mathbf{A}_r) \in \mathbb{R}^{r \times rp}$. We extend each $(r \times r)$-block at the bottom and right with zeroes to get $(n \times n)$-blocks
and overall get an $(n \times np)$-matrix $\mathbf{B}_{\mathbf{z}}(\mathbf{A}_r)^{(n \times np)}$, which can be seen as a candidate
to solve \eqref{proj_fitting_g}. Thus we have
\begin{equation} \label{W-rxrpTonxnp}
 \mathbf{B}_{\mathbf{z}}(\mathbf{A}_r) \quad = \quad \mathbf{I}_r^T \mathbf{B}_{\mathbf{z}}(\mathbf{A}_r)^{(n \times np)} \underline{\mathbf{I}_r}
\end{equation}
which implies that
\begin{equation} \label{W-rxrpTonxnp-implication}
 \pfaerrr{\mathbf{g}_{\mathbf{B}_{\mathbf{z}}(\mathbf{A}_r)}}{\mathbf{A}_r^T \mathbf{z}}{\mathbf{A}_r^T \hist_{\mathbf{z}, p}} \quad = \quad \pfaerrr{\mathbf{I}_r^T \mathbf{g}_{\mathbf{B}_{\mathbf{z}}(\mathbf{A}_r)^{(n \times np)}}}{\mathbf{A}_r^T \mathbf{z}}{\mathbf{A}^T \hist_{\mathbf{z}, p}}
\end{equation}
Since we know that $\mathbf{B}_{\mathbf{z}}(\mathbf{A})$ is an optimal solution of \eqref{proj_fitting_g}, we have
\begin{equation} \label{W-rxrpTonxnp-implication2}
\pfaerrr{\mathbf{I}_r^T \mathbf{g}_{\mathbf{B}_{\mathbf{z}}(\mathbf{A}_r)^{(n \times np)}}}{\mathbf{A}_r^T \mathbf{z}}{\mathbf{A}^T \hist_{\mathbf{z}, p}} \quad \geq \quad \pfaerrr{\mathbf{I}_r^T \mathbf{g}_{\mathbf{B}_{\mathbf{z}}(\mathbf{A})}}{\mathbf{A}_r^T \mathbf{z}}{\mathbf{A}^T \hist_{\mathbf{z}, p}}
\end{equation}
and thus
\begin{equation} \label{W-rxrpTonxnp-implication3}
\pfaerrrb{\mathbf{g}_{\mathbf{A}_r^T \mathbf{z}}^*}{\mathbf{A}_r^T \mathbf{z}}{\mathbf{A}_r^T \hist_{\mathbf{z}, p}} \quad \geq \quad \pfaerrrb{\mathbf{I}_r^T \mathbf{g}_{\mathbf{A}^T \mathbf{z}}^*}{\mathbf{A}_r^T \mathbf{z}}{\mathbf{A}^T \hist_{\mathbf{z}, p}}
\end{equation}
With the projective orthogonal agnosticity criterion \eqref{projectiveOrthogonalAgnosticity}, we can transform the right side of \eqref{W-rxrpTonxnp-implication3} into the right side of \eqref{informationConsistency}
and have formally shown information consistency for model \eqref{pfa-criterionARGeneralMatrix}.
\end{proof}

\subsection{General formulation of PFA} \label{sec:generalPFA}
The criteria in section \ref{sec:criteria} assure that a problem analog to \eqref{pfaAutoRegressiveWhitened} can be relaxed to a tractable problem like \eqref{pfaNonAutoRegressiveWhitened} and that it can be solved like in the corresponding section. Additionally they assure that the theorem in section \ref{sec:proofOfRelaxationGap} holds and that the procedure from \ref{sec:overfitting} is applicable. For any projective orthogonal agnostic and information consistent prediction model, \eqref{pfaAutoRegressiveWhitened} can be relaxed to
\begin{equation} \label{pfaGeneralNonAutoRegressiveWhitened}
\optmin{\mathbf{A} \in \orth(n)} \; \bigav{\; \norm{\mathbf{A}_r^T(\mathbf{z}-\mathbf{g}_{\mathbf{z}}^*(\hist_{\mathbf{z}, p}))}^2 \; }
\end{equation}
which can be solved by diagonalizing $\bigav{\left(\mathbf{z} - \mathbf{g}_{\mathbf{z}}^*(\hist_{\mathbf{z}, p})\right) \left(\mathbf{z} - \mathbf{g}_{\mathbf{z}}^*(\hist_{\mathbf{z}, p})\right)^T}$ and sorting the smallest eigenvalues to the upper left. To extend this generalization to \eqref{pfaNonAutoRegressiveIterated}, a version of \eqref{V} that only uses $\mathbf{g}_{\mathbf{z}}^*$ is needed. However this construction is straight forward, but can generally not be written as a matrix like in \eqref{V}. One uses $\mathbf{g}_{\mathbf{z}}^*$ only for prediction of the first (i.e. the new) components of $\mathbf{\zeta}(t+1)$, while the other components can be copied from $\mathbf{\zeta}(t)$.

\section{Relaxation Gap Theorem} \label{sec:proofOfRelaxationGap}
\begin{thm} \label{sec:thm2}
For any prediction model class $\mathcal{G}$ that is projective orthogonal agnostic and information consistent, the following holds:
\begin{align}
&\text{If} \hphantom{\text{thenad}} \exists \; r \leq n, \hphantom{r,} \!\!\! \mathbf{A}^* \in \orth(n) \colon \err({\mathbf{A}_r^*}^T \mathbf{z}) \hphantom{{\mathbf{A}_r^{(0)}}} \!\! = \quad 0 \label{thm2Prerequisites}\\
&\text{and} \hphantom{\text{Ifthe}} \nexists \; \tilde{r} > r, \hphantom{n,} \!\!\! \mathbf{A}^* \in \orth(n) \colon \err({\mathbf{A}_{\tilde{r}}^*}^T \mathbf{z}) \hphantom{{\mathbf{A}_r^{(0)}}} \vphantom{{\mathbf{A}_r^{(0)}}^T} \!\! = \quad 0 \label{thm2Prerequisites2}\\
&\text{then} \hphantom{\text{Ifad}} \hphantom{\nexists \; \tilde{r} > r, \hphantom{n,} \mathbf{A}^* \in \orth(n) \colon} \!\!\! \err({\mathbf{A}_r^{(0)}}^T \mathbf{z}) \hphantom{{\mathbf{A}_r^*}} \!\! = \quad 0\label{thm2}
\end{align}
\end{thm}
Line \eqref{thm2Prerequisites2} has the purpose to ensure that the maximal $r$ holding \eqref{thm2Prerequisites} is used in line \eqref{thm2}.


To prove the theorem, we need to setup some lemmas and the following definition:
\begin{definition}[Space-partition preserving orthogonal transformations]
\begin{equation}
\orth(r, s) \quad \coloneq \quad \diag(\orth(r), \orth(s))
\end{equation}
\end{definition}
This means that every $\tilde{\mathbf{A}} \in \orth(r, s) \subset \orth(r+s)$ has the form
$\left( \begin{smallmatrix} \mathbf{A}_{rr} & \mathbf{0} \\ \mathbf{0} & \mathbf{A}_{ss} \end{smallmatrix} \right)$
with $\mathbf{A}_{rr} \in \orth(r)$ and $\mathbf{A}_{ss} \in \orth(s)$.
Now we can elegantly formulate the following lemma, which deals with the non-uniqueness of $\mathbf{A}^{(0)}$:
\begin{lem} \label{sec:lem1}
Let $\mathbf{A}^{(0)}$ and $\mathbf{B}^{(0)}$ be two different global solutions of \eqref{pfaGeneralNonAutoRegressiveWhitened} and assume that
the best $r$ components are well defined (i.e. component $r+1$ has worse error than component $r$).
\begin{equation}
\exists \; \tilde{\mathbf{A}} \in \orth(r, n-r) \colon \quad \mathbf{A}^{(0)} = \mathbf{B}^{(0)} \tilde{\mathbf{A}}
\end{equation}
\end{lem}
A problem would arise, if for instance the $r$th-worst component and the $(r+1)$th-worst component had equal error.
In that case it would not be well defined, which signal space to extract and the lemma would not hold.

\begin{proof}[Proof of Lemma \ref{sec:lem1}]
As mentioned earlier, we can obtain one global solution by diagonalizing
$\bigav{\left(\mathbf{z} - \mathbf{g}_{\mathbf{z}}^*(\mathbf{Z})\right) \left(\mathbf{z} - \mathbf{g}_{\mathbf{z}}^*(\mathbf{Z})\right)^T}$
and sorting the $r$ smallest eigenvalues to the upper left. Since lemma~\ref{sec:lem1} requires to have a unique choice of $r$ best components,
every optimal solution must have the same set of eigenvalues in the upper left $(r \times r)$-sub-matrix. Thus we can create every solution
by orthogonally transforming the eigenspace of the $r$ smallest eigenvalues in itself. In an analog way, the eigenspace of the $n-r$ largest eigenvalues may be transformed.
The set of partition preserving orthogonal transformations $\orth(r, n-r)$ is exactly defined to consist of the transformations performing this.
\end{proof}

\begin{lem} \label{sec:lem2}
$\forall \; r \leq n, \mathbf{A} \in \orth(n), \tilde{\mathbf{A}} \in \orth(r, n-r) \colon$
\begin{equation} \label{lem2.eq}
\err(\; \mathbf{I}_r^T\mathbf{A}^T \mathbf{z} \;) \quad
 =\quad \err(\; \mathbf{I}_r^T(\mathbf{A}\tilde{\mathbf{A}})^T \mathbf{z} \;) 
\end{equation}
\end{lem}
Lemma \ref{sec:lem2} implies that solutions of \eqref{pfaAutoRegressiveWhitened} stay solutions, if transformed by any $\tilde{\mathbf{A}}~\in~\orth(r, n-r)$.

\begin{proof}[Proof of Lemma \ref{sec:lem2}]
First observe the following fact:
\begin{equation} \label{lem2Proof.1}
\forall \; \tilde{\mathbf{A}} \in \orth(r, n-r) \colon \quad \exists \; \mathbf{A}_{rr} \in \orth(r) \colon \; \mathbf{I}_r^T \tilde{\mathbf{A}}^T = \mathbf{A}_{rr}^T \mathbf{I}_r^T
\end{equation}
With \eqref{lem2Proof.1} and the orthogonal agnosticity criterion it is straight forward to transform 
the right side of \eqref{lem2.eq} into the left:
\begin{equation}
\vphantom{\eq_{\substack{\eqref{lem2Proof.1}\\ \hphantom{\text{orth.}}}}} \err(\mathbf{I}_r^T \tilde{\mathbf{A}}^T \mathbf{A}^T \mathbf{z}) \quad
\eq_{\substack{\eqref{lem2Proof.1}\\ \hphantom{\text{orth.}}}} \quad \err(\mathbf{A}_{rr}^T \mathbf{I}_r^T \mathbf{A}^T \mathbf{z}) \quad
\eq_{\substack{\text{orth.}\\ \text{agn.}}} \quad \err(\mathbf{I}_r^T \mathbf{A}^T \mathbf{z})
\end{equation}

\end{proof}

Now we are ready to assemble the proof of the relaxation gap theorem:
\begin{proof}[Proof of the relaxation gap theorem]
By condition \eqref{thm2Prerequisites}, we have
\begin{equation}
\err({\mathbf{A}_r^*}^T \mathbf{z}) \quad = \quad \pfaerrr{\mathbf{g}_{{\mathbf{A}_r^*}^T \mathbf{z}}^*}{{\mathbf{A}_r^*}^T \mathbf{z}}{{\mathbf{A}_r^*}^T \mathbf{Z}} \hphantom{\mathbf{I}_r^T ._{{.^*}} {.^*} } \!\!\! = \quad 0
\end{equation}
By information consistency, it follows that
\begin{equation}
\hphantom{\err({\mathbf{A}_r^*}^T \mathbf{z}) \quad = \quad} \pfaerrr{\mathbf{I}_r^T \mathbf{g}_{{\mathbf{A}^*}^T \mathbf{z}}^*}{{\mathbf{A}_r^*}^T \mathbf{z}}{{\mathbf{A}^*}^T \mathbf{Z}} \hphantom{._{{._r^*}} {._r^*} } \!\!\! = \quad 0
\end{equation}
So $\mathbf{A}^*$ is a common global optimum of \eqref{pfaAutoRegressiveWhitened} and \eqref{pfaGeneralNonAutoRegressiveWhitened}.
Since \eqref{thm2Prerequisites2} ensures maximality of $r$, we can apply lemma \ref{sec:lem1} and get:
\begin{equation}
\exists \; \tilde{\mathbf{A}} \in \orth(r, n-r) \colon \quad \mathbf{A}^{(0)} = \, \mathbf{A}^* \tilde{\mathbf{A}}
\end{equation}
Finally by lemma \ref{sec:lem2} we conclude that $\mathbf{A}^{(0)}$ must also be an optimum of \eqref{pfaAutoRegressiveWhitened}.
\end{proof}
By continuity arguments, the implication of theorem \ref{sec:thm2} extends to signals with components of low error -- the lower
the error is, the more precise we can find an optimum of \eqref{pfaAutoRegressiveWhitened} by solving \eqref{pfaGeneralNonAutoRegressiveWhitened}.
Providing bounds for the steepness of this continuous relationship is still an open problem and may be subject of future work.

\section{Future Work}
An important aspect of our future work will be the application of PFA to real world problems. We plan to approach scenarios where SFA is known to produce good results, so we can compare PFA and SFA and get a clearer notion for the differences between the paradigms. On the other hand, we have new scenarios in mind, that specifically would benefit from predictable features. For instance we are working on applications related to robotic navigation and the lane-keeping problem of a simulated car.

As mentioned in some sections of this document, another fork of our future work will be to extend the analytic understanding of the heuristic aspects of the algorithm. This way we also aim to improve our methods to avoid overfitting.

\section*{Acknowledgments}
This work is funded by a grant from the German Research Foundation (Deutsche Forschungsgemeinschaft, DFG) to L. Wiskott (SFB 874, TP B3) and
supported by the German Federal Ministry of
Education and Research within the National Network Computational
Neuroscience - Bernstein Fokus: “Learning behavioral models: From human
experiment to technical assistance”, grant FKZ 01GQ0951.

\pagebreak
\bibliography{MachineLearning,Miscellaneous,Predictability,ReinforcementLearning,SFA,Extern,WiskottGroup}
\bibliographystyle{unsrt}

\pagebreak
\appendix
\section{Appendix}
\subsection{Notation overview} \label{sec:notation}
This section gives an overview of the notation used in this paper.

\begin{tabular}{p{1.1cm}p{9.7cm}ll}
$\mathbf{x}(t)$ & denotes the raw input signal and might only be available for a discrete sequence of $t$'s.\\

$\trph$ & $\coloneq~\{t_0, \ldots, t_k\}$ denotes a discrete time sequence (considered as equidistant with step size normalized to $1$). We usually
refer to $\trph$ as the \textit{training phase}.\\

$\av{\mathbf{s}(t)}_{t \in S}$ & $\coloneq~\frac{1}{\abs{S}} \sum_{t\in S} \mathbf{s}(t)$ denotes the average of some signal $\mathbf{s}$ over a finite set $S$.
For $S~=~\trph$ we just write $\av{\mathbf{s}(t)}_t$ or even $\av{\mathbf{s}}$, if it is obvious, what unbound variable is targeted.\\

$\mathbf{h}(\mathbf{x})$ & denotes the expansion function and usually consists of a set of monomials of low degree.\\

$\mathbf{z}(t)$ & denotes $\mathbf{h}(\mathbf{x}(t))$ after sphering it.\\


$\mathbf{m}(t)$ & denotes the optimized output signal ($\mathbf{m}$ for \textit{model}).\\

$n$ & 
denotes the number of components to be analyzed (after expansion).\\

$r$ & denotes the number of extracted components (“features”).\\

$\mathbf{A}, \mathbf{a}$ & denotes the matrix (or vector if $r=1$) holding the linear composition of the output-signal.
We set $\mathbf{m}(t)~=~\mathbf{A}^T\mathbf{z}(t)$.\\

$\mathbf{a}_i$ & denotes the $i$'th column of $\mathbf{A}$, so we can write $m_i(t)~=~\mathbf{a}_i^T \mathbf{z}(t)$.\\

$\orth(n)$ & $\subset \mathbb{R}^{n \times n}$ denotes the orthogonal group of dimension $n$, i.e. $\forall \; \mathbf{A} \in \orth(n) \colon \; \mathbf{A}\mathbf{A}^T = \mathbf{A}^T\mathbf{A} = \mathbf{I}$\\

$p$ & denotes the number of recent signal-values involved in the prediction. We also call it the \textit{prediction-order}.\\

$\mathbf{I}_{s, r}$ & denotes the $s\times r$ identity matrix ($s$ counting rows, $r$ counting columns).
For $s = r$ this is a usual square identity, while in the non-square case it consists
of a square identity block in the top or left area, filled up with zeroes to fit the given shape.\\

$\mathbf{I}_{r}$ & $\coloneq~\mathbf{I}_{n, r}$\\


$\mathbf{A}_r$ & $\coloneq~\mathbf{A}\mathbf{I}_r$


%
\end{tabular}
\renewcommand{\arraystretch}{1}

We frequently use the $p$-step time-history of a signal $\mathbf{z}$, which we formalize by the following function:
\begin{align} \label{history}
	&\hist_{\mathbf{z}, p, \Delta}(t) \quad \coloneq \quad \sum_{i=1}^p \quad \mathbf{z}(t-i\Delta) \mathbf{e}_i^T \quad \text{with} \quad \mathbf{e}_i \in \mathbb{R}^p\\
	&\hist_{\mathbf{z}, p}(t)_{\hphantom{, \Delta}} \quad \coloneq \quad \hist_{\mathbf{z}, p, 1}(t)
\end{align}
Here $\mathbf{e}_i$ denotes the $i$-th $p$-dimensional euclidean unit vector, which is $1$ at position $i$ and $0$ everywhere else.

Further more we sometimes use the Kronecker product $\otimes$ and the $\mvec$-operator defined as follows:

For matrices $\mathbf{A} \in \mathbb{R}^{m \times n}$ and $\mathbf{B} \in \mathbb{R}^{k \times l}$ and with $a_{ij}$ denoting the entries, $\mathbf{a}_i$ the columns of $\mathbf{A}$:

\begin{equation}
 \mathbf{A} \otimes \mathbf{B} \quad \coloneq \quad \left( \begin{matrix} a_{11}\mathbf{B}& \cdots & a_{1n}\mathbf{B} \\ \vdots & \ddots & \vdots \\ a_{m1} \mathbf{B} & \cdots & a_{mn}\mathbf{B} \end{matrix} \right) \; \in \; \mathbb{R}^{mk \times nl}
\end{equation}

\begin{equation}
\mvec(\mathbf{A}) \quad \coloneq \quad \left( \begin{matrix} \mathbf{a}_1\\ \vdots \\ \mathbf{a}_n \end{matrix} \right) \; \in \; \mathbb{R}^{mn}
\end{equation}

Additionally, we sometimes make use of the following shortcut:
\begin{equation} \label{multiA}
\underline{\mathbf{A}} \quad \coloneq \quad \mathbf{I}_{p, p} \otimes \mathbf{A} \quad = \qquad \underbrace{\!\!\!\!\!\!\left( \begin{matrix} \mathbf{A}&  & \mathbf{0} \\  & \ddots &  \\ \mathbf{0} &  & \mathbf{A} \end{matrix} \right)\!\!\!\!\!\!}_{\text{$p$ times $\mathbf{A}$}} 
\end{equation}

\subsection{Extracting predictable single components} \label{sec:extractIsolated}
In section \ref{sec:pfa} we initially stated a prediction model that always scopes on single components.
This idea was not suitable for PFA because it contradicts the orthogonal agnosticity criterion.
In this section we propose a strategy to extract well predictable single components even though. We begin
by recalling our initial notion of linear auto regressive predictability:
\begin{align}
 \mathbf{a}^T \mathbf{z}(t) \quad \appr^! \;& \quad b_{1} \mathbf{a}^T \mathbf{z}(t-1) + \ldots + b_{p} \mathbf{a}^T \mathbf{z}(t-p) \label{pfa-criterionARScalar}\\
= \;& \quad \mathbf{a}^T \hist_{\mathbf{z}, p}(t) \; \mathbf{b}
\end{align}
It is possible to write this for multiple dimensions by constraining the coefficient-matrices to be diagonal:
\begin{equation} \label{pfa-criterionARDiagMatrix}
 \mathbf{m}(t) \quad \appr^! \quad \mathbf{B}_{1} \mathbf{m}(t-1) + \ldots + \mathbf{B}_{p} \mathbf{m}(t-p) \quad \text{with}\quad \mathbf{B}_i \in \mathbb{R}^{n \times n} \text{, diagonal}
\end{equation}
This model is not orthogonal agnostic, so a different approach than in section \ref{sec:pfa} is needed.
To minimize the least-squares-error of \eqref{pfa-criterionARScalar}, the following optimization
problem needs to be solved:

\begin{align} \label{pfaOrigin}
	\begin{split} 
		\optmin{\mathbf{a} \in \mathbb{R}^n,\;\mathbf{b} \in \mathbb{R}^p} & \av{\left( \mathbf{a}^T (\mathbf{z} - \hist_{\mathbf{z}, p} \mathbf{b}) \right)^2 \;} \\
		\subjectto	& \mathbf{a}^T \av{\mathbf{z}} \hphantom{\mathbf{a} \mathbf{z}^T} \, \quad = \quad 0 \quad \quad \text{(zero mean)}\\
				& \mathbf{a}^T \av{\mathbf{z} \mathbf{z}^T} \mathbf{a} \quad = \quad 1 \quad \quad \text{(unit variance)}
	\end{split}
\end{align}
Via analytic optimization it is straight forward to find the optimal $\mathbf{a}$, if $\mathbf{b}$ is fixed and vice versa:

If $\mathbf{b}$ is fixed, choose $\mathbf{a}$ as the eigenvector corresponding to the smallest eigenvalue in
\begin{equation} \label{pfa_b_to_a}
	\bigav{\mathbf{z}\mathbf{z}^T} - \bigav{\mathbf{z}\mathbf{b}^T\hist_{\mathbf{z}, p}} - \bigav{\hist_{\mathbf{z}, p}^T\mathbf{b}\mathbf{z}^T} + \bigav{\hist_{\mathbf{z}, p}\mathbf{b}\mathbf{b}^T\hist_{\mathbf{z}, p}^T}
\end{equation}

If $\mathbf{a}$ is fixed, choose $\mathbf{b}$ as
\begin{equation} \label{pfa_a_to_b}
	\mathbf{b}^T \; \coloneq \; \bigav{\mathbf{z}^T\mathbf{a}\mathbf{a}^T\hist_{\mathbf{z}, p}} \bigav{\hist_{\mathbf{z}, p}^T\mathbf{a}\mathbf{a}^T\hist_{\mathbf{z}, p}}^{-1}
\end{equation}
By inserting \eqref{pfa_a_to_b} into \eqref{pfaOrigin} one could obtain a problem written in $\mathbf{a}$ only:
\begin{align} \label{pfaOriginConcerning_a}
	\begin{split} 
		\optmin{\mathbf{a} \in \mathbb{R}^n} & \av{\left( \mathbf{a}^T (\mathbf{z} - \hist_{\mathbf{z}, p} \bigav{\mathbf{z}^T\mathbf{a}\mathbf{a}^T\hist_{\mathbf{z}, p}} \bigav{\hist_{\mathbf{z}, p}^T\mathbf{a}\mathbf{a}^T\hist_{\mathbf{z}, p}}^{-1}) \right)^2 \;} \\
		\subjectto	& \mathbf{a}^T \bigav{\mathbf{z}} \hphantom{\mathbf{a} \mathbf{z}^T} \, \quad = \quad 0 \quad \quad \text{(zero mean)}\\
				& \mathbf{a}^T \bigav{\mathbf{z} \mathbf{z}^T} \mathbf{a} \quad = \quad 1 \quad \quad \text{(unit variance)}
	\end{split}
\end{align}
Problem \eqref{pfaOriginConcerning_a} is not efficiently globally solvable by any method known to us, which is mainly due to the occurrence of $\mathbf{a}$ in a matrix-term
under an inversion-symbol. However a possible strategy is to approximate the solution by choosing an initial value for $\mathbf{a}$ or $\mathbf{b}$ and
applying \eqref{pfa_b_to_a} and \eqref{pfa_a_to_b} in turns until a stable state is reached.

As a reasonable initial value for this procedure we choose $\mathbf{b}$ such that it is the best predictor of $\mathbf{z}$ on average, in absence of any $\mathbf{a}$:
\begin{equation} \label{averageBest_b}
	\mathbf{z}(t) \quad \appr^! \quad b_{1} \mathbf{z}(t-1) + \ldots + b_{p} \mathbf{z}(t-p) \quad = \quad \hist_{\mathbf{z}, p}(t) \mathbf{b}
\end{equation}
To minimize the error of \eqref{averageBest_b} on average over all components of $\mathbf{z}$, we propose the following least-squares optimization:

\begin{equation} \label{startB}
	\optmin{\mathbf{b} \in \mathbb{R}^p} \av{ (\mathbf{z} - \hist_{\mathbf{z}, p} \mathbf{b})^T (\mathbf{z} - \hist_{\mathbf{z}, p} \mathbf{b})}
\end{equation}
The solution of this problem is
\begin{equation} \label{startBSolution}
	\mathbf{b} \; \coloneq \; \bigav{\mathbf{z}^T\hist_{\mathbf{z}, p}} \bigav{\hist_{\mathbf{z}, p}^T\hist_{\mathbf{z}, p}}^{-1}
\end{equation}
Solution \eqref{startBSolution} does not change, if we replace $\mathbf{z}$ by $\mathbf{A}^T\mathbf{z}$ with any orthogonal, full ranked $\mathbf{A}$.
However, one quickly finds examples, where the procedure stabilizes in sub-optimal states. Though one can partly overcome this issue by
estimating better starting points, the method still has unknown success-probability.

Probably a better possibility is to solve \eqref{pfaOrigin} with PFA as described in section \ref{sec:extraction} for $r = 1$.

After extracting one component either way, one can project $\mathbf{z}$ to the signal space uncorrelated (i.e.\ orthogonal) to the extracted component. The extraction- and projection-procedure can be repeated until any desired number of components is extracted.

%

\end{document}